\documentclass[letterpaper, 10 pt, conference]{ieeeconf}  
\usepackage{multirow}
\usepackage{makecell}
\usepackage{graphicx} 
\usepackage{booktabs}
\usepackage{amssymb}
\usepackage{amsmath}
\usepackage{url}
\usepackage[hidelinks,breaklinks=true]{hyperref}

\IEEEoverridecommandlockouts                              

\title{\LARGE \bf
TempoBridge: Language-Guided Tempo Control for Vision-Language-Action Policies
}
\author{Yeonseo Lee$^{1}$,~Hyosup Shin$^{1}$,~Guebin Hwang$^{1}$,~and~Sungho Jo$^{1\dagger}$%
\thanks{$^{1}$Korea Advanced Institute of Science and Technology (KAIST), Daejeon, South Korea.}%
\thanks{$^{\dagger}$Corresponding author.}%
}

\begin{document}

\IEEEaftertitletext{%
\parbox{\textwidth}{%
    \makebox[\textwidth][c]{%
        \includegraphics[width=0.95\textwidth]{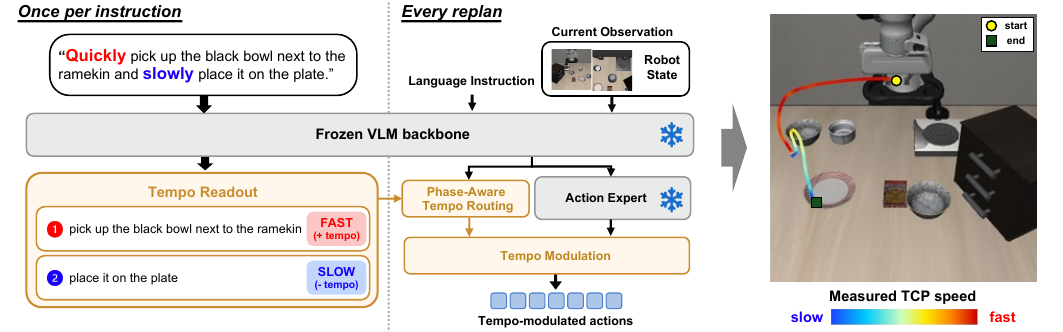}%
    }\\[-1mm]
    \refstepcounter{figure}%
    \label{fig:figure1}%
    {\footnotesize
    Fig.~\thefigure.
    \textbf{Language-conditioned tempo control with TempoBridge.}
    TempoBridge reads tempo cues from the internal representations of a frozen VLA policy and aligns them with task progress to adjust execution speed according to the instruction.
    }%
    \vspace{2mm}%
}%
}

\maketitle
\thispagestyle{empty}
\pagestyle{empty}


\begin{abstract}
Vision-Language-Action (VLA) models are effective at understanding \emph{what} task to perform, but provide limited control over \emph{how} it should be executed, such as moving quickly or slowly.
We introduce TempoBridge, a lightweight framework that uses frozen VLA representations to modulate actions according to tempo cues in the instruction at each task phase, 
without additional tempo-conditioned robot demonstrations or tempo-specific base-policy fine-tuning.
TempoBridge extracts tempo cues from contextual VLM representations,
aligns them with task progress through a causal phase router, and modulates nominal motion commands during execution.
Across LIBERO tasks, TempoBridge improves Tempo Success Rate from 52.6\% to 89.7\% under canonical tempo instructions while retaining high task success.
It also preserves near-baseline performance when no tempo cue is present and generalizes to unseen tempo expressions without additional training.
Experiments on a physical robot further demonstrate language-conditioned tempo modulation in real-world manipulation.
Project page: \url{https://lysees.github.io/tempobridge-page/}
\end{abstract}

\section{INTRODUCTION}
\label{sec:introduction}
Vision-Language-Action (VLA) models have recently emerged as a promising
paradigm for general-purpose robot manipulation, enabling a single policy to
execute diverse tasks from natural-language instructions \cite{zitkovich2023rt2,kim2024openvla,black2025pi0}. 
Existing VLAs increasingly understand what to do, but provide limited control over how to execute it \cite{jing2026tempovla,haon2025mechanistic}.
In many practical settings, however, execution manner is an important part of
the user's intent \cite{shao2025langwbc,belkhale2024rth}.
For example, a robot may be asked to move \emph{quickly} for efficiency or \emph{slowly} for deliberate execution.
These preferences may also vary as a task progresses, as in ``\emph{quickly pick up the object, then slowly place it on the target}.''
In this work, we study this capability as \emph{language-conditioned tempo control}.

A straightforward approach is to train a policy on trajectories associating tempo instructions with execution speeds.
However, collecting demonstrations at different speeds is costly and difficult to scale across tasks, objects, and environments \cite{khazatsky2024droid,o2024open,walke2023bridgedata}.
Temporal augmentation can reduce the cost of collecting additional demonstrations \cite{jing2026tempovla}, but incorporating the augmented trajectories into the policy still requires computationally expensive fine-tuning \cite{kim2025fine}.

Pretrained VLA representations may already encode semantic distinctions associated with manner expressions such as quickly and slowly, and prior work has shown that internal features related to such concepts can be identified and manipulated to influence VLA behavior \cite{haon2025mechanistic}.
This raises the possibility of using semantic tempo information available in the frozen representation as an interface for execution control, rather than learning tempo interpretation entirely from newly collected robot trajectories.
The key challenge is to connect these semantic tempo cues to an explicit execution-time control mechanism while accounting for task progress.
We therefore ask: \emph{Can semantic tempo cues from a frozen VLA be connected to execution-time motion modulation, enabling language-conditioned tempo control without tempo-specific robot demonstrations or policy fine-tuning?}

To address this question, we introduce \textbf{TempoBridge}, a lightweight framework that reads semantic tempo information from a frozen VLA’s internal language representations and connects it to phase-aware motion modulation.
TempoBridge takes a frozen $\pi_{0.5}$ policy \cite{black2025pi05} as its base policy and uses a \emph{prototype-based tempo readout} to extract a compact tempo sequence from the instruction.
When the requested tempo varies within a task, a \emph{causal phase router} aligns this sequence with task progress using visual features reused from the policy and robot-state history.
The selected tempo then modulates the nominal motion commands during closed-loop execution.
The base VLA continues to generate the nominal task actions, 
while TempoBridge modulates their execution according to the requested tempo and estimated task phase.

This design keeps the pretrained policy fixed while separating tempo interpretation and phase estimation from nominal action generation. The tempo readout is constructed from text examples, whereas the phase router is trained on existing task demonstrations annotated with phase-transition intervals, without tempo labels. 
Consequently, TempoBridge requires neither tempo-conditioned robot demonstrations nor tempo-specific fine-tuning of the base policy.
When no tempo cue is detected, the readout returns NORMAL and the nominal actions remain unchanged. 
Because the readout operates on contextual VLM representations, the same fixed readout can also be applied to previously unseen tempo expressions without retraining.

We evaluate TempoBridge on 40 manipulation tasks across the Spatial, Object, Goal, and Long suites of LIBERO \cite{liu2023libero}.
TempoBridge raises Tempo Success Rate from 52.6\% to 89.7\% under canonical quickly/slowly instructions, 
while preserving near-baseline performance without tempo cues and improving tempo compliance for unseen expressions. 
Real-robot experiments further validate language-conditioned tempo modulation.

Our main contributions are:
\begin{itemize}
    \item We formulate language-conditioned tempo control for frozen VLAs and investigate whether semantic tempo cues in their internal representations can serve as an interface for execution-time motion modulation, without additional tempo-conditioned robot demonstrations or tempo-specific base-policy fine-tuning.

    \item We introduce \textbf{TempoBridge}, a lightweight framework that combines prototype-based tempo readout, causal phase routing, and execution-time action modulation while keeping the entire base policy frozen.

    \item We evaluate physical tempo compliance, task competence under original instructions, and transfer to unseen tempo expressions across 40 LIBERO tasks, and further validate TempoBridge in real-world manipulation experiments with a physical robot.
    
\end{itemize}

\section{Related Work}
\label{sec:related_work}

\subsection{Vision-Language-Action Models}
Vision-Language-Action (VLA) models transfer semantic and world knowledge from pretrained vision-language models into robotic control. RT-2 \cite{zitkovich2023rt2} established this paradigm by co-finetuning web-scale VLMs with robot trajectories in a shared token space, and subsequent generalist policies scaled language-conditioned manipulation across heterogeneous embodiments \cite{ghosh2024octo, kim2024openvla} or coupled pretrained representations with continuous action experts for dexterous, long-horizon execution \cite{black2025pi0, black2025pi05}. Across these paradigms, however, language specifies \emph{what} should be achieved, while \emph{how} an action is executed remains implicit in the demonstration distribution. Temporal characteristics such as speed or caution, therefore, do not surface as explicit control variables, even though the corresponding linguistic concepts are plausibly already represented in the pretrained backbone.

\subsection{Controlling Execution in Pretrained VLAs}
Approaches to controlling \emph{how} a policy executes differ mainly in what part of the system they modify, and therefore in what supervision they require.

One line modifies data and weights. Language motions as an intermediate action representation, counterfactual relabeling, and semantically aligned action tokenizers \cite{belkhale2024rth, glossop2025cast, wenjie2026lost} all learn how a linguistic modifier maps to physical execution from robot trajectories. TempoVLA \cite{jing2026tempovla} conditions on an explicit speed scalar and re-times demonstrations by temporal augmentation, with a VLM scheduler dispatching per-chunk speeds at deployment. This still requires retraining the policy, and tempo arrives as a numeric input from an external module.

A second line intervenes on a frozen policy's internal activations, differing in how the directions are identified \cite{haon2025mechanistic, buurmeijer2026observing, khan2025sparse} and how the intervention is regulated once found \cite{babu2026ctrlsteer, khayatan2026dimas}. H\"aon \textit{et al.} \cite{haon2025mechanistic} show that linguistic concepts such as \emph{slow} survive VLA training in the backbone and remain causally linked to end-effector motion, though they identify features heuristically by token similarity and steer by overriding their activations with a selected coefficient rather than a physically interpretable execution variable. Buurmeijer \textit{et al.} \cite{buurmeijer2026observing} make such intervention systematic by training linear observers for robot states and actions and computing the minimal representation-space edit that places an observed feature inside a desired interval. In both cases, the target is an externally specified number rather than read from language, and identifying the directions requires either manual inspection or labeled robot data.
CTRL-STEER \cite{babu2026ctrlsteer} instead regulates the magnitude of the intervention online with a feedback controller, although the target concept is still supplied externally and the controller is fitted per task.

In contrast, TempoBridge reads tempo cues from the language instruction through frozen VLA representations and connects them to phase-aware motion modulation, without tempo-conditioned demonstrations or tempo-specific VLA fine-tuning.
\section{Method}
\begin{figure*}
\centering
\includegraphics[width=2\columnwidth]{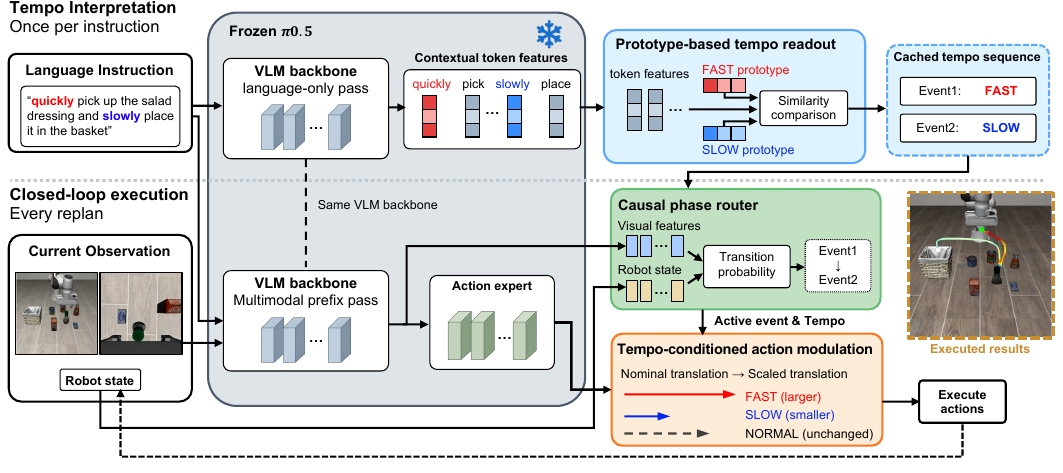}
\caption{Overview of TempoBridge.
Using the frozen VLM backbone within $\pi_{0.5}$, a prototype-based tempo readout extracts an ordered tempo sequence from the instruction once and caches it for execution.
At each replan, a causal phase router uses visual features reused from the policy forward pass and robot state to track task progress and select the active event.
The corresponding tempo coefficient scales the translational components of the nominal action chunk.}
\label{fig:figure2}
\vspace{-10pt}
\end{figure*} 
\subsection{Overview and Problem Formulation}

We introduce \textbf{TempoBridge}, a lightweight framework that interprets language-specified tempo using contextual representations from a frozen vision-language-action (VLA) policy and connects it to phase-aware motion modulation.
As illustrated in Fig.~\ref{fig:figure2}, TempoBridge uses a frozen $\pi_{0.5}$ policy~\cite{black2025pi05} as its base policy, consisting of a PaliGemma~\cite{beyer2024paligemma} VLM backbone and an action expert.
We use \emph{tempo} to denote the qualitative execution pace associated with a task event.
At each replanning step, the base policy produces a nominal action chunk from the current observation and language instruction, while all of its parameters remain fixed.

TempoBridge consists of three components: a prototype-based tempo readout, a causal phase router, and tempo-conditioned action modulation.
The readout processes the instruction once and extracts a cached
\emph{tempo sequence} $z \in \mathcal{Z}$, where
\[
\begin{aligned}
\mathcal{Z} = \{&
[\textsc{Normal}],\,
[\textsc{Fast}],\,
[\textsc{Slow}],\\
&
[\textsc{Fast},\textsc{Slow}],\,
[\textsc{Slow},\textsc{Fast}]
\}.
\end{aligned}
\]
A single-entry sequence specifies one tempo throughout execution.
A two-entry sequence specifies the requested tempo order over two successive task events, for which we assume a single forward transition between events.

During closed-loop execution, the causal phase router tracks task progress and selects the active entry of the cached tempo sequence.
The corresponding tempo is then used to modulate the nominal action chunk. For a single-entry sequence, the router is bypassed and the same tempo is applied throughout the rollout.

The readout and router use different representations from the same frozen VLM
backbone. The readout performs a language-only forward pass once per instruction, whereas the router reuses visual features from the multimodal policy forward pass at every replan.
Prototype construction uses text instructions, 
while router training uses existing task trajectories with phase-transition annotations.
Neither component requires tempo-conditioned robot demonstrations or fine-tuning of the base VLA, and no external language model is used.

\subsection{Prototype-Based Tempo Readout}
\subsubsection{Contextual Features and Tempo Prototypes}
For each instruction, we perform a language-only forward pass through the PaliGemma VLM backbone within the frozen $\pi_{0.5}$ policy.
This pass uses the same VLM parameters as the multimodal policy pass but is executed only once per instruction.
Valid language tokens attend bidirectionally to the instruction, allowing their representations to incorporate the surrounding linguistic context.

Motivated by prior evidence that behaviorally meaningful concepts can be recovered from internal transformer representations of pretrained VLAs~\cite{haon2025mechanistic,buurmeijer2026observing},
we use the residual-stream output after the twelfth transformer block as a fixed contextual representation for tempo readout.
Let $\mathbf{h}_i \in \mathbb{R}^{2048}$ denote the representation of valid language token $i$, excluding padding and boundary tokens, and let $\hat{\mathbf{h}}_i=\mathbf{h}_i/\|\mathbf{h}_i\|_2$ denote its normalized feature.
We construct the readout using 30 task instructions from LIBERO-90 and reserve 10 additional instructions from distinct tasks for validation.
For each training instruction, we prepend \emph{quickly} to obtain a \textsc{Fast} example and \emph{slowly} to obtain a \textsc{Slow} example, while the unmodified instruction serves as a tempo-free example.
This yields 30 contextual cue features for each tempo class and 30 tempo-free examples.
Only the training examples are used to construct the prototypes and compute the readout thresholds.

Let $\mathcal{C}_F$ and $\mathcal{C}_S$ denote the sets of canonical cue-token indices for FAST and SLOW, respectively.
For $c\in\{F,S\}$, we construct a unit-norm prototype \cite{snell2017prototypical}

\begin{equation}
\mathbf{p}_{c}
=
\operatorname{normalize}
\left(
\frac{1}{|\mathcal{C}_{c}|}
\sum_{i\in\mathcal{C}_{c}}
\hat{\mathbf{h}}_{i}
\right).
\label{eq:tempo_prototype}
\end{equation}

Thus, prototype construction requires only normalization and averaging and does not involve gradient-based readout training.
For each token, we compute its cosine similarity to prototype $c$ as $s_i^c=\hat{\mathbf{h}}_i^\top\mathbf{p}_c$ \cite{vinyals2016matching}.

To distinguish tempo cues from ordinary instruction tokens, we calibrate one detection threshold per class.
Let $\mathcal{N}$ contain all valid token features from the 30 tempo-free instructions.
The threshold is defined as the midpoint between the mean similarity of the canonical cue features and that of the tempo-free tokens:

\begin{equation}
\tau_c
=
\frac{1}{2}
\left(
\frac{1}{|\mathcal{C}_c|}
\sum_{i\in\mathcal{C}_c}s_i^c
+
\frac{1}{|\mathcal{N}|}
\sum_{i\in\mathcal{N}}s_i^c
\right).
\label{eq:tempo_threshold}
\end{equation}

The tempo-free term is computed over individual tokens rather than sentence-level averages.
The prototypes and thresholds are fixed before evaluation.

\subsubsection{Token-Level Tempo Classification}
Each token is classified by combining its absolute similarity to the prototypes with competition between the FAST and SLOW classes:

\begin{equation}
y_i =
\begin{cases}
\textsc{Fast},
& s_i^F > \tau_F \ \land\ s_i^F > s_i^S, \\[2pt]
\textsc{Slow},
& s_i^S > \tau_S \ \land\ s_i^S > s_i^F, \\[2pt]
\textsc{Other},
& \text{otherwise}.
\end{cases}
\label{eq:tempo_token_decision}
\end{equation}

Here, \textsc{Other} indicates that a token is not selected as a tempo cue and is distinct from \textsc{Normal}, which denotes the default execution tempo.

Because a word may be split into multiple subtokens, conflicting FAST and SLOW predictions can occur within the same word.
We group subtokens using tokenizer surface and offset information. 
When a word contains both labels, we sum the threshold excess $s_i^c-\tau_c$ over the subtokens assigned to each class and assign the entire word to the class with the larger total excess.
Ties are assigned \textsc{Other}. 
Words without a FAST-SLOW conflict retain their preliminary labels.

\subsubsection{Tempo Sequence Extraction}
After word-level correction, we identify contiguous regions of FAST and SLOW
tokens.
For each tempo class, we retain the region with the largest mean threshold
excess.
If both classes are present, the order of the selected region midpoints determines whether the decoded sequence is
$[\textsc{Fast},\textsc{Slow}]$ or
$[\textsc{Slow},\textsc{Fast}]$.
If only one class is present, the sequence contains that tempo alone; 
if neither class is detected, the readout returns $[\textsc{Normal}]$.

The readout therefore decodes the requested order of tempo expressions rather than explicitly assigning each modifier to a particular verb or action.
For multi-tempo instructions, the causal phase router determines which entry of the decoded sequence is active during execution.
\subsection{Causal Phase Routing}
When the readout returns a two-entry tempo sequence, TempoBridge must determine which task event is currently active.
The causal phase router performs this alignment online without access to oracle task-phase information at deployment.
It maintains recurrent state across replanning steps and permits only a single forward transition from Event~1 to Event~2.
For single-entry sequences, the router is not used.
\subsubsection{Reused Visual Features and State History}
At replanning step $r$, we obtain a visual feature $\mathbf{v}_r\in\mathbb{R}^{2048}$ by mean-pooling the image-token outputs from the multimodal prefix pass already used for action generation.
The router therefore does not require an additional visual-backbone forward pass.
These visual features are distinct from the language-only features used by the tempo readout.

We project $\mathbf{v}_r$ to a 32-dimensional PCA representation $\bar{\mathbf{v}}_r$ and combine it with the robot state $\mathbf{q}_r\in\mathbb{R}^{8}$, which contains end-effector position, rotation, and gripper state.
The router input is
\[
\mathbf{x}_r =
[
\bar{\mathbf{v}}_r,\,
\Delta\bar{\mathbf{v}}_r,\,
\mathbf{q}_r,\,
\Delta\mathbf{q}_r
],
\]
where
$\Delta\bar{\mathbf{v}}_r=\bar{\mathbf{v}}_r-\bar{\mathbf{v}}_{r-1}$ and
$\Delta\mathbf{q}_r=\mathbf{q}_r-\mathbf{q}_{r-1}$.
Both differences are initialized to zero at $r=0$.
The PCA projection and component-wise standardization statistics are fitted on the router training data and fixed thereafter.

\subsubsection{Causal Transition Estimation}
A GRU processes the standardized router inputs and predicts a transition hazard $h_r$ \cite{cho2014learning}.
The hazards are accumulated into a monotonic transition probability \cite{kvamme2021continuous}:

\begin{align}
\mathbf{u}_r
&=
\operatorname{GRU}_{\phi}
\left(
\operatorname{standardize}(\mathbf{x}_r),
\mathbf{u}_{r-1}
\right), \\
h_r
&=
\sigma(\mathbf{w}^{\top}\mathbf{u}_r+b),
\qquad
F_r
=
1-\prod_{j=0}^{r}(1-h_j).
\label{eq:phase_router}
\end{align}

Execution begins in Event~1 and switches to Event~2 when $F_r \geq 0.5$.
The transition is one-way, so Event~2 remains active for the remainder of the rollout.

\subsubsection{Interval-Supervised Router Training}
We train the router on 40 existing demonstrations from 20 LIBERO-90 tasks and validate it on 20 demonstrations from 10 additional tasks, using two demonstrations per task.
The selected tasks cover object transfer, drawer manipulation, stove operation, and bowl stacking.
Each demonstration is annotated with an interval containing the transition between the first and second task events.
Transition intervals are first proposed automatically and then manually reviewed and corrected when necessary.

We use intervals rather than single transition timestamps because the semantic boundary between two manipulation events can be ambiguous.
Let $\mathcal{D}_{\mathrm{phase}}$ denote the annotated router-training set.
For each training trajectory $n$, let $(L_n,R_n]$ denote the annotated interval containing the transition.
Because $F_r$ is the cumulative transition probability, the probability that the transition occurs within this interval is $F_{R_n}-F_{L_n}$ \cite{turnbull1976empirical}.
We therefore optimize

\begin{equation}
\mathcal{L}_{\mathrm{phase}}
=
-\frac{1}{|\mathcal{D}_{\mathrm{phase}}|}
\sum_{n\in\mathcal{D}_{\mathrm{phase}}}
\log
\left(
F_{R_n}^{(n)}-F_{L_n}^{(n)}
\right).
\label{eq:phase_loss}
\end{equation}

Only the router parameters are optimized, while the VLA feature extractor remains frozen.
The router is trained exclusively on the 40 training demonstrations and remains fixed during deployment.

\subsection{Tempo-Conditioned Action Modulation}
At each replanning step, the tempo associated with the active event determines a modulation gain

\begin{equation}
\alpha(z)=
\begin{cases}
1.3, & z=\textsc{Fast},\\
1.0, & z=\textsc{Normal},\\
0.7, & z=\textsc{Slow}.
\end{cases}
\label{eq:tempo_gain}
\end{equation}
We use fixed heuristic gains symmetric around 1.0, without task-specific tuning, to test whether language-derived tempo cues induce consistent physical speed differences.
For a single-entry tempo sequence, the same gain is used throughout execution.
For a two-entry sequence, the gain changes when the phase router switches the active event.

For the LIBERO experiments, we instantiate tempo-conditioned modulation by scaling the Cartesian translation components of the nominal action chunk.
Let $\Delta\mathbf{p}_{r,j}\in\mathbb{R}^{3}$ denote the translational component of action $j$ at replanning step $r$, and let $\mathbf{b}_{r,j}$ contain the remaining rotation and gripper commands.
TempoBridge applies

\begin{equation}
\tilde{\mathbf{a}}_{r,j}
=
\left[
\alpha(z_{e_r})\Delta\mathbf{p}_{r,j},
\mathbf{b}_{r,j}
\right],
\qquad
j=0,\ldots,H-1,
\label{eq:tempo_modulation}
\end{equation}

where $e_r$ denotes the active event.
The gain is selected once per replan and applied uniformly to the predicted action chunk.

The robot executes the first $M\leq H$ actions of the modulated chunk,
receives a new observation, and replans.
Thus, in the LIBERO setting, TempoBridge changes the magnitude of the nominal translation commands while retaining the policy's predicted rotation and gripper commands.
The gain acts on commanded translation rather than directly prescribing physical TCP speed.
The realized motion instead emerges from the closed-loop interaction between action modulation, execution, and subsequent replanning.

\section{Experiments}

We evaluate TempoBridge along four complementary axes.
First, we test whether language-specified tempo can be realized in physical execution while preserving the underlying task competence.
Second, we evaluate whether TempoBridge leaves the base policy largely
unchanged when no tempo cue is present.
Third, we test generalization to tempo expressions that are never used to construct or configure the readout.
Finally, we analyze the individual contributions of language interpretation,
phase routing, and motion realization, and validate the resulting tempo control on a physical robot.
\begin{table*}[t!]
\centering
\caption{Task, tempo, and task \& tempo success rates (\%) under
\emph{quickly}/\emph{slowly} instructions.}
\label{tab:table1}

\setlength{\tabcolsep}{4.5pt}
\renewcommand{\arraystretch}{1.05}
\small

\resizebox{\textwidth}{!}{%
\begin{tabular}{l cc cc cc cc ccc}
\toprule
\multirow{2}{*}{\textbf{Method}}
& \multicolumn{2}{c}{\textbf{Spatial}}
& \multicolumn{2}{c}{\textbf{Object}}
& \multicolumn{2}{c}{\textbf{Goal}}
& \multicolumn{2}{c}{\textbf{Long}}
& \multicolumn{3}{c}{\textbf{Overall}} \\
\cmidrule(lr){2-3}
\cmidrule(lr){4-5}
\cmidrule(lr){6-7}
\cmidrule(lr){8-9}
\cmidrule(lr){10-12}

& \makecell{\textbf{Task}\\\textbf{Success Rate} $\uparrow$}
& \makecell{\textbf{Tempo}\\\textbf{Success Rate} $\uparrow$}
& \makecell{\textbf{Task}\\\textbf{Success Rate} $\uparrow$}
& \makecell{\textbf{Tempo}\\\textbf{Success Rate} $\uparrow$}
& \makecell{\textbf{Task}\\\textbf{Success Rate} $\uparrow$}
& \makecell{\textbf{Tempo}\\\textbf{Success Rate} $\uparrow$}
& \makecell{\textbf{Task}\\\textbf{Success Rate} $\uparrow$}
& \makecell{\textbf{Tempo}\\\textbf{Success Rate} $\uparrow$}
& \makecell{\textbf{Task}\\\textbf{Success Rate} $\uparrow$}
& \makecell{\textbf{Tempo}\\\textbf{Success Rate} $\uparrow$}
& \makecell{\textbf{Task \& Tempo}\\\textbf{Success Rate} $\uparrow$} \\
\midrule

$\pi_{0.5}$
& 94.0 & 48.9 
& 96.0 & 49.5
& \textbf{93.0} & 57.8
& \textbf{91.5} & 54.4
& \textbf{93.6} & 52.6 & 48.8 \\

\textbf{TempoBridge}
& \textbf{94.5} & \textbf{91.0}
& \textbf{97.5} & \textbf{90.8}
& 92.5 & \textbf{93.8}
& 87.0 & \textbf{82.9}
& 92.9 & \textbf{89.7} & \textbf{82.0} \\

\bottomrule
\end{tabular}%
}
\end{table*}
\begin{figure*}
\centering
\includegraphics[width=2\columnwidth]{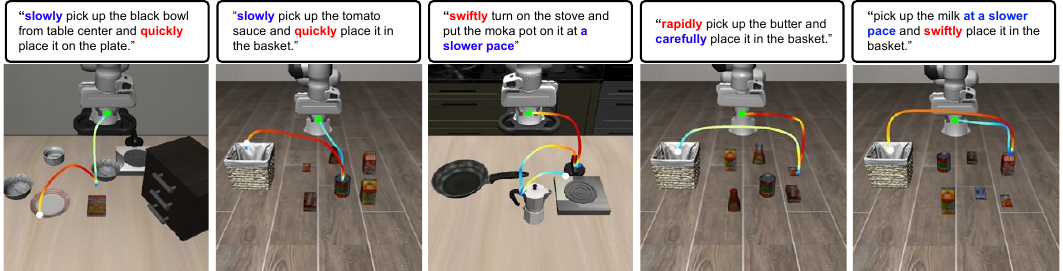}
\caption{Qualitative examples of TempoBridge execution on LIBERO.
The left two panels use \emph{quickly}/\emph{slowly} cues, while the right three use tempo expressions unseen during readout construction.
Each panel shows the instruction and executed TCP trajectory, colored by measured TCP speed from slower motion (blue) to faster motion (red).}
\label{fig:figure3}
\vspace{-10pt}
\end{figure*} 
\subsection{Experimental Setup and Evaluation Metrics}
We evaluate on 40 tasks from the four LIBERO \cite{liu2023libero} suites: Spatial, Object, Goal, and Long, with 10 tasks per suite.
Both methods use the same $\pi_{0.5}$ \cite{black2025pi05} checkpoint trained for LIBERO.
The baseline directly executes the nominal policy actions, 
whereas TempoBridge augments the same policy with the prototype-based tempo readout, causal phase router, and tempo-conditioned action modulation.
The two methods use identical instructions and matched initial states and random seeds.

The tempo prototypes are constructed only from task-context sentences containing \emph{quickly} and \emph{slowly}. 
The phase router is trained exclusively on existing demonstrations from LIBERO-90 tasks that are disjoint from all 40 evaluation tasks, using task-transition annotations but no tempo labels. 
No tempo-conditioned robot demonstrations are collected, and all $\pi_{0.5}$ parameters remain frozen.
For LIBERO, TempoBridge scales the Cartesian translation components of the nominal action by gains of 1.3, 1.0, and 0.7 for \textsc{Fast}, \textsc{Normal}, and \textsc{Slow}, respectively.

\noindent\textbf{Task Success Rate.}
Task Success Rate measures the fraction of all rollouts satisfying the standard LIBERO task-success criterion, independently of tempo compliance.

\noindent\textbf{Tempo Success Rate.}
Tempo Success Rate measures whether a successfully executed task realizes the requested relative tempo.
We compute translational tool-center-point (TCP) speed from consecutive positions.
For each evaluated segment, we discard the first and last 25\% of speed samples in temporal order and average the central 50\% to reduce boundary effects.
For two-event instructions, evaluation boundaries are obtained from simulator-state predicates independently of the phase router.
A rollout satisfies the tempo instruction when the mean TCP speed of the \textsc{Fast} event exceeds that of the \textsc{Slow} event.

For single-event instructions, \textsc{Fast} and \textsc{Slow} correspond to separate rollouts.
We compare each task-successful rollout against task-successful rollouts with the opposite tempo under matched initial states and random seeds.
When multiple opposite-tempo lexical variants are available, their mean successful speed is used as the reference.
Tempo Success Rate is therefore computed over task-successful rollouts for which the required tempo comparison is available.
It should be interpreted as a conditional measure of physical tempo realization after successful task execution.

\noindent\textbf{Cue Recognition Accuracy.}
For unseen tempo expressions, we separately evaluate whether the readout assigns the expected tempo label.
For two-event instructions, each position of the predicted tempo sequence is evaluated against its corresponding cue independently of the companion cue.
For a single-event instruction, the readout must return the expected single tempo label.
This metric separates readout errors from downstream execution errors.


\subsection{Tempo Control under Canonical Instructions}

We first evaluate explicit tempo control using the canonical \emph{quickly} and \emph{slowly} expressions.
For 27 two-event tasks, we construct both \textsc{Fast}$\rightarrow$\textsc{Slow} and
\textsc{Slow}$\rightarrow$\textsc{Fast} instructions.
For the remaining 13 single-event tasks, we construct separate \textsc{Fast} and \textsc{Slow} instructions.
This yields 80 instructions, each evaluated over 10 rollouts, for a total of 800 rollouts per method.

Table~\ref{tab:table1} shows that TempoBridge increases the overall Tempo Success Rate from 52.6\% to 89.7\%, while the Task Success Rate remains comparable to the baseline (92.9\% vs.\ 93.6\%).
Because Tempo Success Rate is evaluated only among task-successful, tempo-evaluable rollouts, we additionally report Task \& Tempo Success Rate, defined over all rollouts.
Under this metric, TempoBridge improves from 48.8\% to 82.0\%, showing that the gain in tempo compliance is retained when task completion and tempo compliance are evaluated jointly.

\begin{table}[!t]
\centering
\caption{Task success rates (\%) on LIBERO benchmark instructions without tempo cues.}
\label{tab:table2}

\fontsize{7}{8}\selectfont
\setlength{\tabcolsep}{3.5pt}
\renewcommand{\arraystretch}{1.05}

\begin{tabular*}{\columnwidth}{@{\extracolsep{\fill}}lccccc@{}}
\toprule
\textbf{Method}
& \textbf{Spatial}
& \textbf{Object}
& \textbf{Goal}
& \textbf{Long}
& \textbf{Overall} \\
\midrule

$\pi_{0.5}$
& 95.0
& \textbf{97.0}
& \textbf{95.0}
& \textbf{93.0}
& \textbf{95.0} \\

\textbf{\textbf{TempoBridge}}
& \textbf{98.0}
& 96.0
& \textbf{95.0}
& 88.0
& 94.3 \\

\bottomrule
\end{tabular*}
\end{table}
\begin{table*}[!t]
\centering
\caption{Task, tempo, and task \& tempo success rates (\%) for tempo expressions unseen during readout construction.}
\label{tab:table3}

\setlength{\tabcolsep}{4.5pt}
\renewcommand{\arraystretch}{1.05}
\small

\resizebox{\textwidth}{!}{%
\begin{tabular}{l cc cc cc cc ccc}
\toprule
\multirow{2}{*}{\textbf{Method}}
& \multicolumn{2}{c}{\textbf{Spatial}}
& \multicolumn{2}{c}{\textbf{Object}}
& \multicolumn{2}{c}{\textbf{Goal}}
& \multicolumn{2}{c}{\textbf{Long}}
& \multicolumn{3}{c}{\textbf{Overall}} \\
\cmidrule(lr){2-3}
\cmidrule(lr){4-5}
\cmidrule(lr){6-7}
\cmidrule(lr){8-9}
\cmidrule(lr){10-12}

& \makecell{\textbf{Task}\\\textbf{Success Rate} $\uparrow$}
& \makecell{\textbf{Tempo}\\\textbf{Success Rate} $\uparrow$}
& \makecell{\textbf{Task}\\\textbf{Success Rate} $\uparrow$}
& \makecell{\textbf{Tempo}\\\textbf{Success Rate} $\uparrow$}
& \makecell{\textbf{Task}\\\textbf{Success Rate} $\uparrow$}
& \makecell{\textbf{Tempo}\\\textbf{Success Rate} $\uparrow$}
& \makecell{\textbf{Task}\\\textbf{Success Rate} $\uparrow$}
& \makecell{\textbf{Tempo}\\\textbf{Success Rate} $\uparrow$}
& \makecell{\textbf{Task}\\\textbf{Success Rate} $\uparrow$}
& \makecell{\textbf{Tempo}\\\textbf{Success Rate} $\uparrow$}
& \makecell{\textbf{Task \& Tempo}\\\textbf{Success Rate} $\uparrow$} \\
\midrule

$\pi_{0.5}$
& \textbf{95.6} & 50.3 
& \textbf{99.3} & 49.5
& \textbf{96.1} & 61.2
& \textbf{89.5} & 53.1
& \textbf{95.1} & 53.5 & 50.6 \\

\textbf{TempoBridge}
& 94.8 & \textbf{79.7}
& 96.1 & \textbf{75.2}
& 93.8 & \textbf{91.0}
& 84.8 & \textbf{74.2}
& 92.3 & \textbf{80.1} & \textbf{72.9} \\

\bottomrule
\end{tabular}%
}
\vspace{-10pt}
\end{table*}

\subsection{Task Competence without Tempo Cues}
%
We next evaluate whether TempoBridge preserves the underlying policy behavior when the instruction contains no explicit tempo cue.
We use the 40 original LIBERO instructions without added tempo expressions and evaluate 10 rollouts per instruction.
The tempo readout returns \textsc{Normal} for 38 of the 40 instructions (95\%), leaving the nominal translation commands unchanged for those instructions.
As shown in Table~\ref{tab:table2}, TempoBridge achieves an overall Task Success Rate of 94.3\%, compared with 95.0\% for the baseline.
These results show that TempoBridge largely preserves the task competence of the underlying policy when no tempo modification is requested.

\subsection{Generalization to Unseen Tempo Expressions}

\begin{table}[!t]
\centering
\caption{Cue-specific tempo recognition accuracy (\%) for expressions unseen during readout construction.}
\label{tab:table4}

\fontsize{7}{8.5}\selectfont
\setlength{\tabcolsep}{3.5pt}
\renewcommand{\arraystretch}{1.05}

\begin{tabular*}{\columnwidth}{@{\extracolsep{\fill}}lcc@{}}
\toprule
\textbf{Tempo expression}
& \textbf{Target tempo}
& \textbf{Accuracy $\uparrow$} \\
\midrule

rapidly
& FAST
& 98.3 \\

swiftly
& FAST
& 98.3 \\

carefully
& SLOW
& 57.9 \\

at a slower pace
& SLOW
& 95.0 \\

\bottomrule
\end{tabular*}
\end{table}
Because TempoBridge reads tempo cues from contextual representations of the frozen VLM rather than relying on a fixed lexical rule set, we test whether the readout can transfer to tempo expressions that are not used during its construction or configuration.
We use \emph{rapidly} and \emph{swiftly} as \textsc{Fast} cues, and \emph{carefully} and \emph{at a slower pace} as \textsc{Slow} cues.
All TempoBridge components remain fixed throughout this evaluation.
For each two-event task, we combine each \textsc{Fast} expression with each \textsc{Slow} expression and assign them to the two task clauses in both \textsc{Fast}$\rightarrow$\textsc{Slow} and \textsc{Slow}$\rightarrow$\textsc{Fast} orders, yielding eight instructions per task.
For each single-event task, we prepend each of the four unseen expressions to the benchmark instruction.
This produces 216 two-event instructions evaluated over 10 rollouts each and 52 single-event instructions evaluated over 20 rollouts each, for a total of 3,200 rollouts per method.

As shown in Table~\ref{tab:table3}, TempoBridge increases Tempo Success Rate from 53.5\% to 80.1\% while maintaining a Task Success Rate of 92.3\%.
The Task \& Tempo Success Rate also improves from 50.6\% to 72.9\%. 
Although tempo performance is lower than in the canonical setting, these results show that TempoBridge retains substantial tempo control for expressions that were never used to construct the readout.

To better understand this degradation, Table~\ref{tab:table4} reports cue-level recognition accuracy.
The readout recognizes \emph{rapidly} and \emph{swiftly} as \textsc{Fast} with 98.3\% accuracy each, and \emph{at a slower pace} as \textsc{Slow} with 95.0\% accuracy, whereas accuracy for \emph{carefully} drops to 57.9\%.
The high recognition rates for expressions with explicit tempo semantics indicate that the frozen VLM representations support transfer beyond the canonical cues used for readout construction, while the lower accuracy for \emph{carefully} suggests that generalization is more challenging when the relation to execution speed is semantically indirect.
Representative TempoBridge executions with unseen expressions are shown in Fig.~\ref{fig:figure3}, while additional qualitative comparisons with the $\pi_{0.5}$ baseline are provided on the project page.

\subsection{Component-wise Analysis}
\begin{table}[!t]
\centering
\caption{Component-wise diagnostics of TempoBridge under canonical and unseen tempo expressions.}
\label{tab:table5}

\fontsize{7}{8}\selectfont
\setlength{\tabcolsep}{2.0pt}
\renewcommand{\arraystretch}{1.10}

\begin{tabular*}{\columnwidth}{@{\extracolsep{\fill}}lccc@{}}
\toprule
\textbf{Setting}
& \makecell{\textbf{Readout}\\\textbf{Acc.} $\uparrow$}
& \makecell{\textbf{Balanced Phase}\\\textbf{Agreement} $\uparrow$}
& \makecell{\textbf{Median Speed}\\\textbf{Ratio}} \\
\midrule

Canonical
& 97.5\%
& 89.7\%
& 1.58$\times$ \\

Unseen
& 78.4\%
& 90.4\%
& 1.52$\times$ \\

\bottomrule
\end{tabular*}
\end{table}
We separately evaluate the tempo readout and phase router, and quantify the magnitude of the resulting physical speed separation.
Readout Accuracy measures exact agreement between the decoded and requested tempo sequences over unique evaluation instructions. 
For the phase router, we compute the fraction of replanning steps correctly assigned within each reference event and average the two event-wise scores, giving both events equal weight.
This Balanced Phase Agreement is then averaged over task-successful two-event rollouts for which the router is activated and reference segmentation is available.
The reference segmentation is defined independently of the router, and rollouts in which the router never transitions remain included.
Tempo Success Rate only requires the measured \textsc{Fast} speed to exceed the \textsc{Slow} speed, regardless of the magnitude of the difference. 
We therefore additionally report the median $\bar{v}_{\mathrm{Fast}}/\bar{v}_{\mathrm{Slow}}$ ratio over the same task-successful comparisons used for Tempo Success Rate, with each mean speed computed over the central 50\% of its evaluation segment.

As shown in Table~\ref{tab:table5}, Readout Accuracy is 97.5\% and 78.4\% under canonical and unseen-expression instructions, respectively, while Balanced Phase Agreement reaches 89.7\% and 90.4\%.
The corresponding median \textsc{Fast}-to-\textsc{Slow} speed ratios are $1.58\times$ and $1.52\times$.
These results verify the operation of the readout and phase router while showing that the realized speed differences are substantial rather than marginally satisfying the binary tempo criterion.

\subsection{Real-World Tempo Control}
\begin{figure}
\centering
\includegraphics[width=\columnwidth]{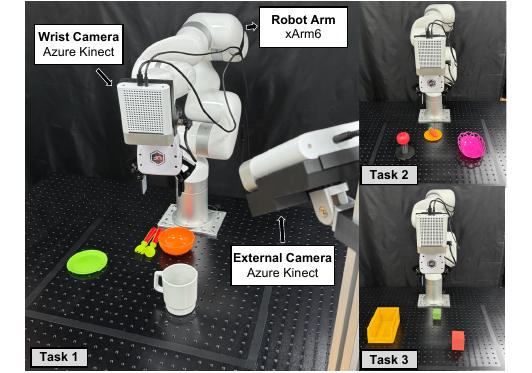}
\caption{Real-world experimental setup with an xArm6 robot, external and wrist-mounted Azure Kinect cameras, and the three manipulation task configurations.}
\label{fig:figure4}
\vspace{-10pt}
\end{figure}

We finally evaluate whether TempoBridge can use language-specified tempo to produce measurable changes in physical execution speed using the real-robot setup shown in Fig.~\ref{fig:figure4}.
We consider three manipulation tasks:
Task~1, ``Pick up the white mug and place it on the green coaster'';
Task~2, ``Pick up the red apple and place it in the pink basket''; and
Task~3, ``Pick up the red block and drop it into the yellow box.''

We compare the task-adapted $\pi_{0.5}$ baseline and TempoBridge under \textsc{Fast}$\rightarrow$\textsc{Slow} (FS) and \ \textsc{Slow}$\rightarrow$\textsc{Fast} (SF) instructions, with \emph{quickly} and \emph{slowly} assigned to the two task clauses in the corresponding order.
Both methods use the same task-specific checkpoints and execution settings without tempo-specific policy fine-tuning.

Across all real-world trials, TempoBridge achieved 60 successful executions out of 67 attempts (89.6\%), while the task-adapted $\pi_{0.5}$ baseline achieved 60 out of 66 (90.9\%).
We analyze 10 task-successful rollouts per method, task, and tempo ordering, for 120 rollouts in total.
For real-world control, the tempo gain scales joint displacements and motion limits.
We compute translational TCP speed over two segments, from execution start to gripper close and from gripper close to release. These boundaries are independent of the phase router; for each segment, we report duration-weighted mean speed over its central 50\%.
\begin{figure}
\centering
\includegraphics[width=\columnwidth]{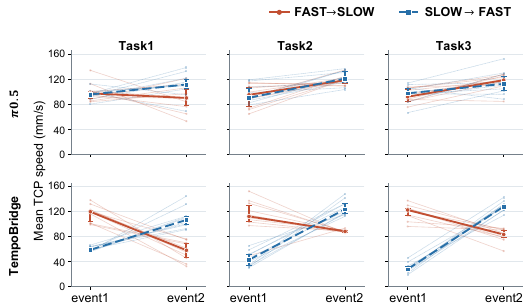}
\caption{Real-world TCP speed under \textsc{Fast}$\rightarrow$\textsc{Slow} (FS) and \textsc{Slow}$\rightarrow$\textsc{Fast} (SF) instructions.
Columns correspond to the three tasks, with the task-adapted $\pi_{0.5}$ baseline in the top row and TempoBridge in the bottom row.
Thin lines represent individual task-successful rollouts, connecting the mean TCP speeds of Event~1 and Event~2.
Bold lines indicate the median, and error bars show the interquartile range over 10 successful rollouts per condition.
}
\label{fig:figure5}
\vspace{-10pt}
\end{figure} 
As shown in Fig.~\ref{fig:figure5}, TempoBridge produces higher median TCP speed for segments instructed with \emph{quickly} than with \emph{slowly} across all three tasks. Swapping the cue order reverses the speed ordering, whereas the task-adapted $\pi_{0.5}$ baseline shows smaller and less consistent changes. These results demonstrate language-conditioned physical tempo control without speed-conditioned demonstrations or tempo-specific policy fine-tuning.

\section{CONCLUSIONS}

We presented TempoBridge, a lightweight framework for making language-specified execution tempo actionable in a frozen VLA.
Rather than learning a tempo-conditioned policy from additional speed-conditioned robot data, TempoBridge leverages linguistic structure available in the pretrained VLM representation and connects it to phase-aware motion modulation.
Across 40 LIBERO tasks, the method substantially improves tempo compliance while largely preserving task competence, maintains near-baseline performance when no tempo cue is present, and generalizes to previously unseen tempo expressions.
Real-world experiments further demonstrate that language-specified tempo can be used to drive measurable changes in execution speed on a physical robot. 
These results suggest that pretrained VLA representations can provide a useful interface not only for specifying \emph{what} a robot should do, but also for controlling aspects of \emph{how} the behavior is executed.

TempoBridge currently focuses on discrete FAST/SLOW tempo control and relatively simple task-phase structures.
Future work could extend the framework to continuous tempo control, more complex multi-stage tasks, and broader execution attributes beyond speed.
More generally, our results motivate further study of how semantic information available in pretrained VLA representations can be used as an interface to explicit execution-time control mechanisms without modifying the underlying policy.


\bibliographystyle{IEEEtran}
\bibliography{IEEEabrv,ref}

\end{document}